\documentclass[journal]{IEEEtran}

\usepackage{algorithm}
\usepackage{algorithmic}
\usepackage{lscape}
\usepackage{makecell}
\usepackage{threeparttable}
\usepackage{tabularray}
\usepackage{float}
\usepackage{amsmath}
\usepackage{amsfonts}
\usepackage{graphicx}
\usepackage{multirow}
\usepackage{pifont}
\usepackage[colorlinks,linkcolor=blue]{hyperref}

\ifCLASSINFOpdf
\else
\fi
\begin{document}
%
\title{DAOS: A Multi-modality In-cabin Action Understanding with Driver Action-Object Synergy Dataset}
%
%
%

\author{Yiming~Li\textsuperscript{*}, 
        Dan~Lin\textsuperscript{*},
        Wenfei~Liang,
        Tianyi~Liu,
        Chen~Cai,
        Wenqian~Wang,
        Kim-Hui~Yap,
        Liang~Zhao,
        and~Bingbing~Li
\thanks{*Yiming Li and Dan Lin contributed equally to this work.}
\thanks{Yiming Li, Wenfei Liang, Tianyi Liu, Chen Cai and Kim-Hui Yap are with the School of Electrical and Electronic Engineering, Nanyang Technological University, Singapore. email: yiming008@e.ntu.edu.sg, e190210@e.ntu.edu.sg, liut0038@e.ntu.edu.sg, wenfei001@e.ntu.edu.sg, ekhyap@ntu.edu.sg. \textit{(Corresponding author: Kim-Hui Yap)}}
\thanks{Dan Lin is with the College of Computer Science and Technology, Harbin Engineering University. email: danlin@hrbeu.edu.cn. \textit{(Corresponding author: Dan Lin)}}
\thanks{Liang Zhao is with School of Software, Dalian University of Technology. email: liangzhao@dlut.edu.cn.}
\thanks{Wenqian Wang is with Singapore University of Technology and Design (SUTD), Singapore. email: wenqian\_wang@sutd.edu.sg.}
\thanks{Bingbing Li is with the Continental Automotive Singapore Pte. Ltd. email: bingbing.li@continental-corporation.com.}}

%
%

\markboth{Journal of \LaTeX\ Class Files,~Vol.~14, No.~8, August~2015}%
{Shell \MakeLowercase{\textit{et al.}}: Bare Demo of IEEEtran.cls for IEEE Journals}
%



\maketitle

\begin{abstract}
Multimodal action understanding requires complementary sensory observations to be connected with structured semantic knowledge. This is particularly challenging in vehicle cabins, where fine-grained actions exhibit similar upper-body motion, relevant objects may be small or occluded, and appearance varies substantially across illumination conditions. Existing in-cabin datasets, however, rarely provide synchronized visual modalities together with dense object annotations and explicit action--object alignment, limiting the study of object-grounded multimodal recognition.
To address this gap, we introduce the \underline{D}river \underline{A}ction with \underline{O}bject \underline{S}ynergy (DAOS) dataset, containing 9,787 video clips, 36 fine-grained actions, 15 object categories, and more than 2.5 million annotated object instances. DAOS provides synchronized RGB, infrared, and depth observations from four camera views, thereby combining complementary appearance, low-light, and geometric information. Building on DAOS, we propose the Action-Object-Relation Network (AOR-Net), a vision--language framework that organizes multimodal understanding at three semantic levels: action, object, and human--object relation. Its Chain-of-action Prompting mechanism progressively grounds global motion with localized object and relation cues, while a textual prototype bank supplies corresponding semantic descriptions. A Mixture of Thoughts module then dynamically selects and integrates the most informative semantic levels and visual modalities under object-rich and object-scarce conditions.
Extensive experiments show that AOR-Net consistently outperforms strong baselines on DAOS, demonstrating the effectiveness of combining heterogeneous visual sensing with structured action--object--relation semantics for fine-grained multimodal video understanding.
\end{abstract}


%
\IEEEpeerreviewmaketitle

\section{Introduction}
Multimodal action understanding aims to recognize human behavior by integrating complementary observations and semantic knowledge rather than relying on a single visual stream. 
Different sensing modalities describe different aspects of an event: RGB captures appearance, infrared (IR) remains informative under weak illumination, and depth preserves geometric structure, while language provides semantic anchors for interpreting what a person is doing and how the person interacts with surrounding objects. 
Learning shared representations across RGB, depth, skeleton, and accelerometer observations has demonstrated that heterogeneous sensors can provide complementary evidence for action recognition \cite{elmadany2019multimodal}. 
Within this broader multimedia setting, the vehicle cabin is a representative and demanding environment because driver actions are fine-grained, object-dependent, and frequently affected by occlusion and illumination changes. 
Reliable recognition therefore requires not only temporal motion modeling, but also coordinated reasoning across visual modalities and across action, object, and relation semantics.

\begin{figure}[t]
\centering
\includegraphics[width=\linewidth]{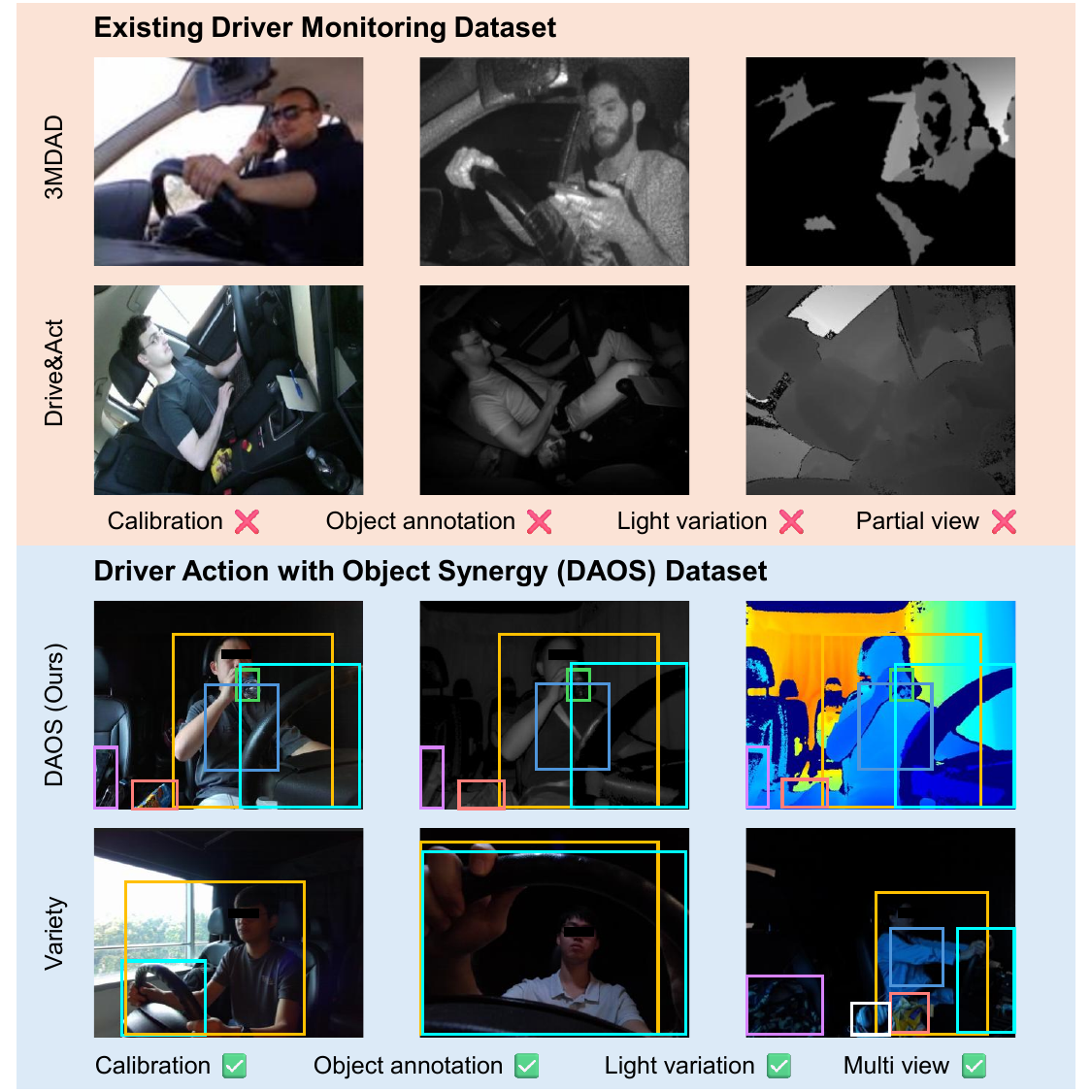}
\caption{Comparison of existing driver-monitoring datasets and DAOS. DAOS provides precise object annotations, multimodal calibration, varied lighting conditions, and multi-view videos for all three visual modalities.}\vspace{-6pt}
\label{fig:motivation}
\end{figure}

\begin{figure*}[t]
\centering
\includegraphics[width=\linewidth]{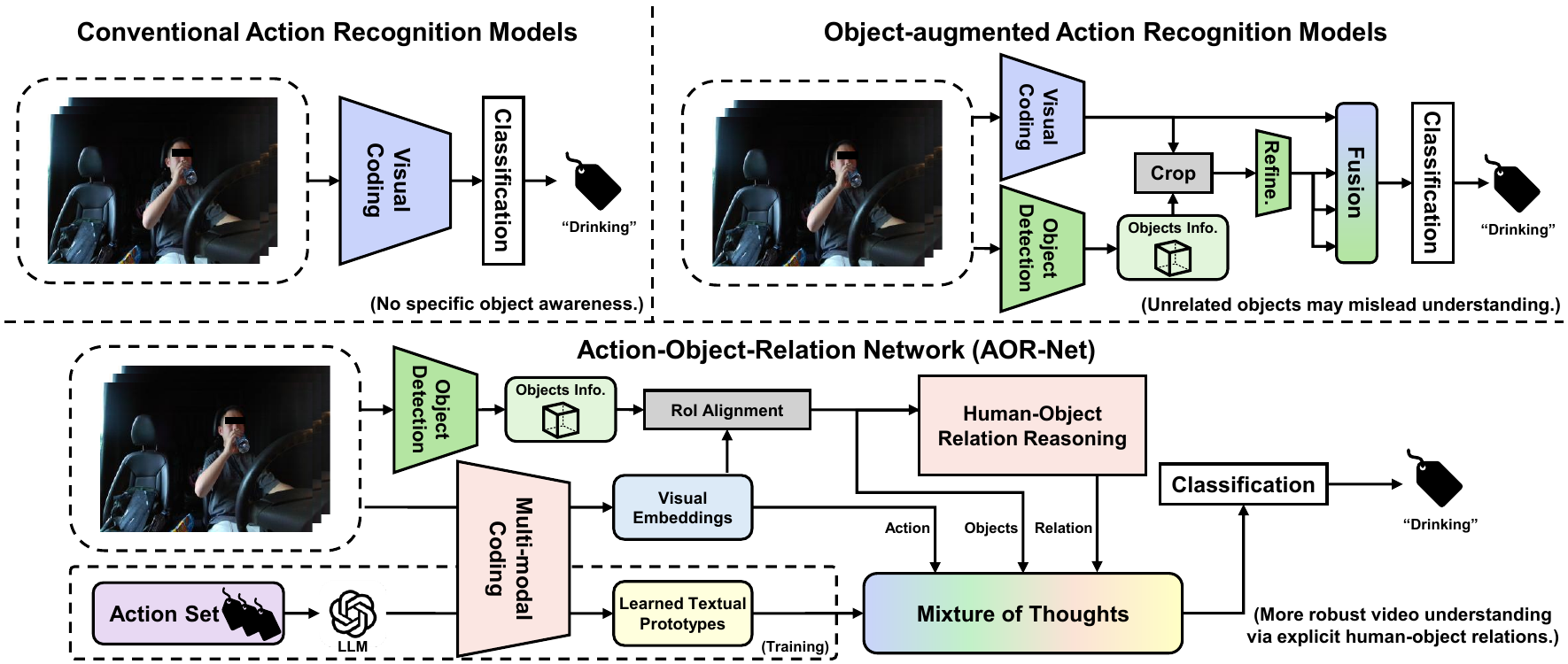}
\caption{
Comparison of model architectures for driver action recognition. 
\textbf{(Top left)} Traditional action recognition models classify actions from global video features without explicit object awareness. 
\textbf{(Top right)} Object-augmented models incorporate bounding boxes or region features but typically fuse them without explicitly determining their semantic relevance. 
\textbf{(Bottom)} The proposed \textbf{Action--Object--Relation Network (AOR-Net)} progressively models action, object, and human--object relation cues and dynamically combines them through the Mixture-of-Thoughts (MoT) module.
}\label{fig:overall}
\vspace{-10pt}
\end{figure*}

Existing driver-monitoring datasets have progressively increased the scale and diversity of in-cabin observations. 
Early datasets, such as DrivFace \cite{diaz2016reduced} and Pandora \cite{borghi2017poseidon}, mainly focus on facial expressions and head poses, whereas Drive\&Act \cite{martin2019drive}, DMD \cite{ortega2022challenges}, 3MDAD \cite{jegham2020novel}, MDM \cite{jha2021multimodal}, and 100-Driver \cite{wang2023100} provide broader action coverage, multiple viewpoints, or multiple sensing modalities. 
These resources have advanced driver behavior analysis, but visually similar actions remain difficult to distinguish in the confined cabin. 
Most movements involve only the upper body, especially subtle hand and arm motions, so global appearance or posture alone may provide nearly identical evidence for actions with different meanings, such as holding a phone, operating a vehicle control, eating, or drinking.
Object-aware recognition can reduce this ambiguity by jointly inferring actions and interacted-object locations from body cues \cite{xu2023skeleton}, or by integrating object detection with interaction reasoning \cite{li2025simultaneous}. 
However, applying such models to in-cabin behavior requires localized object evidence that is temporally aligned with fine-grained action labels.

These challenges reveal two complementary forms of multimodality. 
At the sensory level, RGB, IR, and depth respond differently to lighting, texture, and occlusion, making them suitable for robust observation under varied cabin conditions. 
At the semantic level, action, object, and relation cues answer different questions: global action features describe the overall motion, object features identify the entities involved, and relation features explain how the driver interacts with those entities. 
Text descriptions further provide a common semantic space in which these three levels can be aligned with visual evidence; motion-aware visual and textual prompting has also shown the value of such language guidance for supervised and zero-shot action recognition \cite{gao2025mipclip}. 
Nevertheless, category-level language alignment alone does not explicitly ground semantic descriptions in localized objects or their relations to the human. 
For example, similar hand trajectories may correspond to using a phone or drinking, but the localized object and its relation to the driver's hand or face resolve the ambiguity. 
Thus, in-cabin recognition is not merely a three-stream fusion problem; it requires structured multimodal reasoning that associates heterogeneous visual observations with compositional semantics.
Beyond improving recognition accuracy, this formulation also supports a more interpretable account of model decisions. 
Predictions can be traced from a coarse motion pattern to the object that disambiguates it and then to the relation that explains the driver's intent. 
Such structured evidence is especially useful when one modality becomes unreliable: depth can preserve geometry in darkness, IR can retain body contours, and textual prototypes can stabilize the semantic interpretation of incomplete visual cues.

Current datasets do not fully support this problem formulation. 
Although Drive\&Act \cite{martin2019drive} and DMD \cite{ortega2022challenges} have advanced in-cabin action research, they lack dense frame-level object localization aligned with driver actions (Fig.~\ref{fig:motivation}). 
Retrofitting existing benchmarks is difficult because the RGB, IR, and depth streams in datasets such as Drive\&Act and 3MDAD \cite{jegham2020novel} are not fully aligned, making exhaustive cross-modal object annotation expensive and redundant. 
Moreover, many everyday interaction objects, including cigarettes, headphones, child seats, and face masks, are absent from existing label spaces. 
The resulting gap limits both object-grounded action recognition and systematic analysis of how multiple visual modalities complement one another.

To close the alignment and coverage gaps, we present DAOS, a synchronized multimodal, four-view in-cabin dataset with exhaustive object annotations associated with driver actions. 
DAOS contains 36 fine-grained action classes, 15 object categories, 9,787 action clips, and more than 2.5 million object instances over 74 hours of recordings. 
RGB, IR, and depth data are captured from front, face, left, and right views, providing complementary appearance, low-light, and geometric information. 
The precise temporal correspondence among modalities, actions, and objects makes DAOS a benchmark for multimodal action understanding as well as for action-related human--object interaction analysis.

Building on DAOS, we propose the Action-Object-Relation Network (AOR-Net), a vision--language framework for fine-grained action recognition. 
AOR-Net uses Open-VCLIP \cite{pmlr-v202-weng23b} as its backbone and introduces two components. 
First, the \textbf{Chain-of-action Prompting Framework} progressively constructs action-, object-, and relation-level representations, grounding global video features with localized object regions and human--object interactions. 
The corresponding textual prototype bank encodes action descriptions, object names, and relation descriptions, allowing each visual level to be aligned with an appropriate semantic target. 
Second, the \textbf{Mixture of Thoughts (MoT) Module} dynamically weights the three semantic levels and fuses the outputs of RGB, IR, and depth branches. 
This design adapts to actions for which objects or relations are decisive while retaining robust global evidence when objects are scarce or partially occluded.

We summarize our contributions as follows:
\begin{itemize}
\item We introduce DAOS, a multimodal and multi-view driver-action dataset that synchronizes RGB, IR, and depth observations with fine-grained actions and dense object annotations.
\item We develop AOR-Net, which unifies visual and textual information through progressive action--object--relation prompting and dynamically integrates complementary semantic and sensor cues.
\item Extensive experiments establish strong single- and multimodal baselines on DAOS and demonstrate the effectiveness of object-grounded, relation-aware multimodal action understanding.
\end{itemize}
\section{Related Works}\label{related_works}
\begin{table*}
\centering
\renewcommand\arraystretch{1.2}
\Large
\caption{Comparison of vision-based driver monitoring datasets.}
\label{tab:data_attribute}
\begin{threeparttable}
\resizebox{\linewidth}{!}{
\begin{tabular}{clcccccccccccccccc} 
\hline\hline
\multicolumn{2}{c}{Attributes}                                                                               & \multirow{2}{*}{Year} & \multirow{2}{*}{Size\tnote{a}} & \multirow{2}{*}{Subjects} & \multirow{2}{*}{F/M\tnote{b}} & \multicolumn{2}{c}{Action Annotation} & \multicolumn{3}{c}{Data Modality} & \multicolumn{2}{c}{\begin{tabular}[c]{@{}c@{}}Lighting\\ Conditions\end{tabular}} & \multirow{2}{*}{\begin{tabular}[c]{@{}c@{}}Camera\\ Views\end{tabular}} & \multicolumn{4}{c}{Object annotation}                                                          \\ 
\cline{1-2}\cline{7-13}\cline{15-18}
Scope                                                                                           & Dataset    &                       &                       &                           &                      & Description\tnote{c} & Classes                 & RGB      & IR       & Depth       & Day  & Night                                                                      &                                                                         & Category & Format                                                     & Alignment & Instances  \\ 
\hline
\multirow{5}{*}{\begin{tabular}[c]{@{}c@{}}Driving-related \\ Datasets\end{tabular}}            & DrivFace \cite{diaz2016reduced}   & 2016                  & 0.6k                  & 4                         & 2/2                  & \ding{55}          & \ding{55}                      & 640$\times$480   & \ding{55}       & \ding{55}          & N.A. & N.A.                                                                       & 1                                                                       & \ding{55}       & \ding{55}                                                         & \ding{55}        & \ding{55}         \\
    & DROZY \cite{massoz2016ulg}      & 2016                  & 7h                    & 14                        & 11/3                 & \ding{55}          & \ding{55}                      & \ding{55}       & 512$\times$424   & \ding{55}          & N.A. & N.A.                                                                       & 1                                                                       & \ding{55}       & \ding{55}                                                         & \ding{55}        & \ding{55}         \\
    & Pandora \cite{borghi2017poseidon}    & 2017                  & 250k                  & 22                        & 10/12                & \ding{55}          & \ding{55}                      & 1920$\times$1080 & \ding{55}       & 512$\times$424      & \ding{51}   & N.A.                                                                       & 1                                                                       & \ding{55}       & \ding{55}                                                         & \ding{55}        & \ding{55}         \\
    & DriveAHead \cite{schwarz2017driveahead} & 2017                  & 10.5h                 & 20                        & 4/16                 & \ding{55}          & \ding{55}                      & \ding{55}       & 512$\times$424   & 512$\times$424      & \ding{51}   & N.A.                                                                       & 1                                                                       & \ding{55}       & \ding{55}                                                         & \ding{55}        & \ding{55}         \\
    & DD-Pose \cite{roth2019dd}    & 2019                  & 6h                    & 27                        & 6/21                 & \ding{55}          & 13                      & 2048$\times$2048 & 2048$\times$2048 & \ding{55}          & \ding{51}   & N.A.                                                                       & 2                                                                       & \ding{55}       & \ding{55}                                                         & \ding{55}        & \ding{55}         \\ 
\hline
\multirow{9}{*}{\begin{tabular}[c]{@{}c@{}}Driver Action\\ Recognition \\Datasets\end{tabular}} & SEU \cite{Zhao2012RecognitionOD}        & 2014                  & 29k                   & 20                        & 10/10                & \ding{55}          & 5                       & 640$\times$480   & \ding{55}       & \ding{55}          & \ding{51}   & N.A.                                                                       & 1                                                                       & \ding{55}       & \ding{55}                                                         & \ding{55}        & \ding{55}         \\
    & HEH \cite{ohn2014hand}        & 2014                  & N.A.                  & 8                         & 1/7                  & \ding{55}          & 19                      & 640$\times$480   & \ding{55}       & 640$\times$480      & \ding{51}   & N.A.                                                                       & 1                                                                       & \ding{55}       & \ding{55}                                                         & \ding{55}        & \ding{55}         \\
    & Brain4Cars \cite{jain2016brain4cars} & 2015                  & 2M                    & 10                        & N.A.                 & \ding{55}          & 5                       & 1920$\times$1080 & \ding{55}       & \ding{55}          & \ding{51}   & N.A.                                                                       & 1                                                                       & \ding{55}       & \ding{55}                                                         & \ding{55}        & \ding{55}         \\
    & NTHU-DDD \cite{weng2017driver}   & 2017                  & 210k                  & 36                        & 18/18                & \ding{55}          & 6                       & \ding{55}       & 640$\times$480   & \ding{55}          & \ding{51}   & \ding{51}                                                                         & 1                                                                       & \ding{55}       & \ding{55}                                                         & \ding{55}        & \ding{55}         \\
    & AUC-DD \cite{abouelnaga2017real}     & 2018                  & 144k                  & 44                        & 15/29                & \ding{55}          & 10                      & 1920$\times$1080 & \ding{55}       & \ding{55}          & N.A. & N.A.                                                                       & 1                                                                       & \ding{55}       & \ding{55}                                                         & \ding{55}        & \ding{55}         \\
    & 3MDAD \cite{jegham2020novel}      & 2019                  & 24h                   & 50                        & 12/38                 & \ding{55}          & 16                   & 640$\times$480   & \ding{55} & 640$\times$480      & \ding{51}   & \ding{51}                               & 2         & \ding{55}       & \ding{55}           & \ding{55}        & \ding{55}   \\
    & Drive\&Act \cite{martin2019drive}    & 2019                  & 12h                   & 15                        & 4/11                 & \ding{55}          & 12+34                   & 950$\times$540   & 1280$\times$1024 & 512$\times$424      & \ding{51}   & N.A.                                                                       & 6                                                                       & 17       & label                                                      & \ding{55}        & \ding{55}         \\
    & DMD \cite{ortega2022challenges}        & 2020                  & 41h                   & 37                        & 10/27                & \ding{55}          & 13                      & 1920$\times$1080 & 1280$\times$720  & 1280$\times$720     & \ding{51}   & \ding{51}                                                                         & 3                                                                       & 7        & label                                                      & \ding{55}        & \ding{55}         \\
    & DriverMVT \cite{DriverMVT}  & 2022                  & 36h                   & 9                         & 7/2                  & \ding{55}          & 3                       & 1920$\times$1080 & \ding{55}       & \ding{55}          & \ding{51}   & \ding{51}                                                                         & 2                                                                       & \ding{55}       & \ding{55}                                                         & \ding{55}        & \ding{55}         \\ \cline{2-18}
    & \textbf{DAOS (Ours)} & \textbf{2024}                  & \textbf{74h}                   & \textbf{44}                        & \textbf{10/34}                 & \ding{51}          & \textbf{12+36}                   & \textbf{1920}$\times$\textbf{1080}  & \textbf{1024}$\times$\textbf{1024} & \textbf{1024}$\times$\textbf{1024}    & \ding{51}   & \ding{51}                                                                         & 4                                                                       & \textbf{15}       & \begin{tabular}[c]{@{}c@{}}\textbf{label} +\\ \textbf{location}\end{tabular} & \ding{51}        & \textbf{2.58M}      \\
\hline\hline
\end{tabular}
}
\begin{tablenotes}
\small
\item[N.A.: information not provided by the authors. \quad a: k and M refer to thousand and million frame numbers, h stands for hours in time. \quad b: F for female, M for male. \quad c: Long description of actions and relations.]
\end{tablenotes}
\end{threeparttable}
\vspace{-8pt}
\end{table*}

\subsection{Driver Monitoring Dataset}
In recent years, with advancements in autonomous driving technology and driver monitoring systems, the demand for accurate driver action recognition has been rapidly increasing. To support these research needs, numerous action recognition datasets have been proposed \cite{diaz2016reduced,massoz2016ulg,borghi2017poseidon,martin2019drive,ortega2022challenges,DriverMVT,jha2021multimodal,wang2023100}. Early driving-related datasets, such as DrivFace \cite{diaz2016reduced}, DROZY \cite{massoz2016ulg}, and Pandora \cite{borghi2017poseidon}, addressed certain aspects of in-car research needs but also had limitations. DrivFace \cite{diaz2016reduced} primarily focuses on facial information, DROZY \cite{massoz2016ulg} on drowsiness detection, and Pandora \cite{borghi2017poseidon} records head and shoulder poses. These datasets are limited in size, resolution, and scope, often focusing on a single body part, making it difficult to capture complex actions in the driving scenario.

As demands for accuracy and generalization in driver action recognition have increased, recent datasets, such as Drive\&Act \cite{martin2019drive}, DMD \cite{ortega2022challenges}, DriverMVT \cite{DriverMVT}, MDM \cite{jha2021multimodal}, and 100-Driver \cite{wang2023100}, have introduced higher resolution, larger sizes, and a more comprehensive focus on action recognition or driver attention analysis. For instance, the Drive\&Act \cite{martin2019drive} dataset includes a wide range of fine-grained action annotations in driving-related scenarios, with data captured from five synchronized infrared (IR) views, and a multimodal camera with RGB, depth, and IR. Similarly, the DMD \cite{ortega2022challenges} dataset provides multi-level action annotations across various driving contexts, and the DriverMVT \cite{DriverMVT} dataset also demonstrates significant improvements in resolution and multi-modal data collection. Moreover, it includes heartbeat detection, adding an important physiological dimension to monitor the state of driver. The MDM dataset \cite{jha2021multimodal} further emphasizes naturalistic driver attention analysis with RGB, depth, and CAN-Bus signals, while 100-Driver \cite{wang2023100} highlights large-scale distracted driver classification under cross-view, cross-modality, and cross-vehicle settings. However, while these datasets include object annotations or auxiliary labels, they often only label object categories or temporal intervals (e.g., Drive\&Act \cite{martin2019drive}) and still lack precise object location annotations or dense temporal alignment with actions. This limitation restricts their utility for research on object-augmented action recognition.

To address these issues, we present a new dataset, DAOS (Driver Action with Object Synergy) dataset, aimed at filling the current gaps in the driver monitoring field. DAOS not only increases the duration and the number of participating subjects but also offers a diverse range of action and object categories. DAOS’s primary advantage lies in its precise temporal alignment of actions and objects, as well as detailed location annotations for each object, which make it better suited for supporting research on object-augmented action recognition. With DAOS, we aim to introduce new opportunities and challenges in driver behavior analysis and safety research, further advancing developments in autonomous driving and driver monitoring technology. A comparison of dataset attributes is shown in Table~\ref{tab:data_attribute}

\vspace{-8pt}
\subsection{Video-based Action Recognition and Understanding}
Video-based action recognition has evolved rapidly from early convolutional models to recent transformer-based architectures that capture long-range temporal dependencies. 
Classical approaches such as Two-Stream CNNs \cite{simonyan2014two} and 3D ConvNets (e.g., C3D \cite{tran2015learning}, I3D \cite{carreira2017quo}) model spatial and temporal cues jointly but suffer from limited temporal receptive fields. 
To improve efficiency, Temporal Shift Module (TSM) \cite{lin2019tsm} introduces channel shifting operations to enable temporal modeling within 2D CNNs, achieving strong performance with lightweight computation. 
Similarly, Temporal Relation Networks (TRN) \cite{zhou2018temporal} and SlowFast Networks \cite{feichtenhofer2019slowfast} enhance temporal reasoning by modeling multi-scale motion dynamics across frames.

With the rise of transformers, vision-based action understanding entered a new phase. 
TimeSformer \cite{bertasius2021timesformer} applies divided space-time attention to video tokens, enabling scalable temporal reasoning with high interpretability. 
ViViT \cite{arnab2021vivit} further extends the standard Vision Transformer (ViT) to videos through factorized temporal--spatial attention, improving representation capacity for large-scale benchmarks. 
Mformer \cite{patrick2021keeping} and Uniformer \cite{li2022uniformer} integrate CNN inductive biases into transformer blocks to balance local and global modeling. 
Recent self-supervised methods, such as VideoMAE \cite{tong2022videomae}, leverage masked reconstruction to learn spatio-temporal representations without labels, offering strong pretraining benefits for downstream recognition and retrieval tasks.

Beyond classification, video action understanding research has expanded toward temporal grounding, reasoning, and compositional recognition. 
Methods like ActionCLIP \cite{wang2021actionclip} and X-CLIP \cite{ni2022expanding} bridge vision and language via contrastive pretraining, enhancing generalization to open-vocabulary and zero-shot settings. 
More recently, Open-VCLIP \cite{pmlr-v202-weng23b} adapts vision-language models to video-level reasoning, providing a strong multimodal backbone for context-aware action recognition.

Despite the impressive progress, most existing methods are designed for general-purpose video understanding and do not explicitly account for the constrained, object-centric nature of in-cabin scenes. 
They often rely on large-scale motion cues or scene transitions, which are absent in driver monitoring where subtle hand-object relations dominate. 
Meanwhile, transportation-oriented driver monitoring studies have shown that gaze estimation, attention tracking, and distraction recognition require robust modeling under realistic in-vehicle conditions \cite{vicente2015driver,mittal2023cat}. 
This motivates our design of DAOS and AOR-Net, which emphasize fine-grained human--object relations and temporally aligned multi-modal fusion tailored to real-world in-cabin systems.

\vspace{-5pt}
\subsection{Object-augmented Action Recognition} \label{object_augment}
Since objects in actions play a key role in determining action categories \cite{li2024open}, there is an increasing trend to integrate object information into action-related tasks, such as action recognition and action detection \cite{elsayed2022savi++,locatello2020object,wu2021towards}. 

The field of object-augmented action recognition has been advanced by several innovative models. STLT \cite{radevski2021revisiting} advocates for the fusion of layout and appearance-based models using multi-head attention. STIN \cite{materzynska2020something} explores geometric relations between objects and subjects to enhance compositional action recognition, proving effective in generalization tasks. STRG \cite{wang2018videos} utilizes space-time region graphs and Graph Convolutional Networks, leading to notable performance gains in action recognition.

With the development of ViT, ORViT \cite{herzig2022object} integrates object-centric representations within transformer layers, using dual streams to process object appearance and dynamics, which has demonstrated significant improvements across various recognition tasks. ObjectViViT \cite{zhou2023can} employs an object-guided token sampling and attention module, optimizing performance while reducing computational load. Lastly, ObjectLearner \cite{zhang2022object} introduces a transformer model that learns object-centric summary vectors for videos, showing superior transferability to new tasks and unseen scenarios. These models collectively push the boundaries of incorporating object-level information for human action recognition. However, most of these methods do not consider the relations between certain objects and actions, which leads to the injection of unrelated object information.
\vspace{-2pt}
\section{Dataset}\label{Dataset}

\begin{figure*}[h]
\centering
\includegraphics[width=\textwidth]{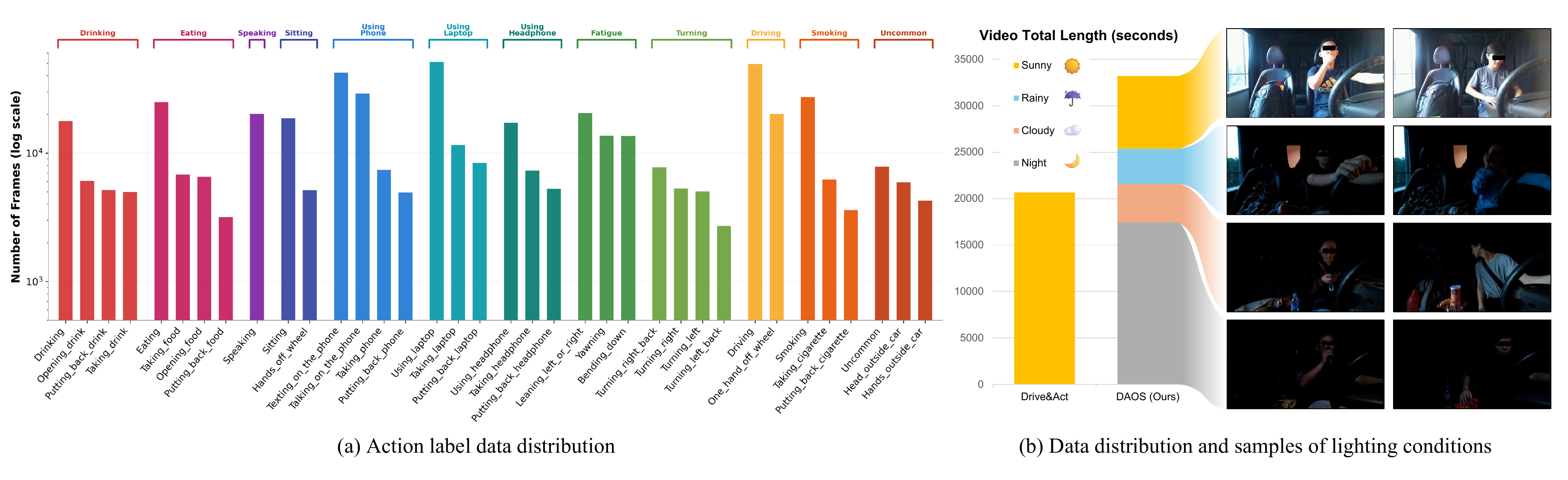}\vspace{-5pt}
\caption{Statistics of our proposed DAOS dataset. (a) Action label data distribution on a logarithmic scale: at the coarse level, consecutive actions involving the same object are merged into 12 major categories; at the fine‑grained level, those are split into 36 more detailed action classes.  
(b) Data distribution comparison of lighting conditions for existing and our proposed dataset. Image samples for different lighting conditions, covering sunny, rainy, cloudy and night conditions. }
\label{fig:category}\vspace{-2pt}
\end{figure*}

\begin{figure*}[h]
\centering
\includegraphics[width=\textwidth]{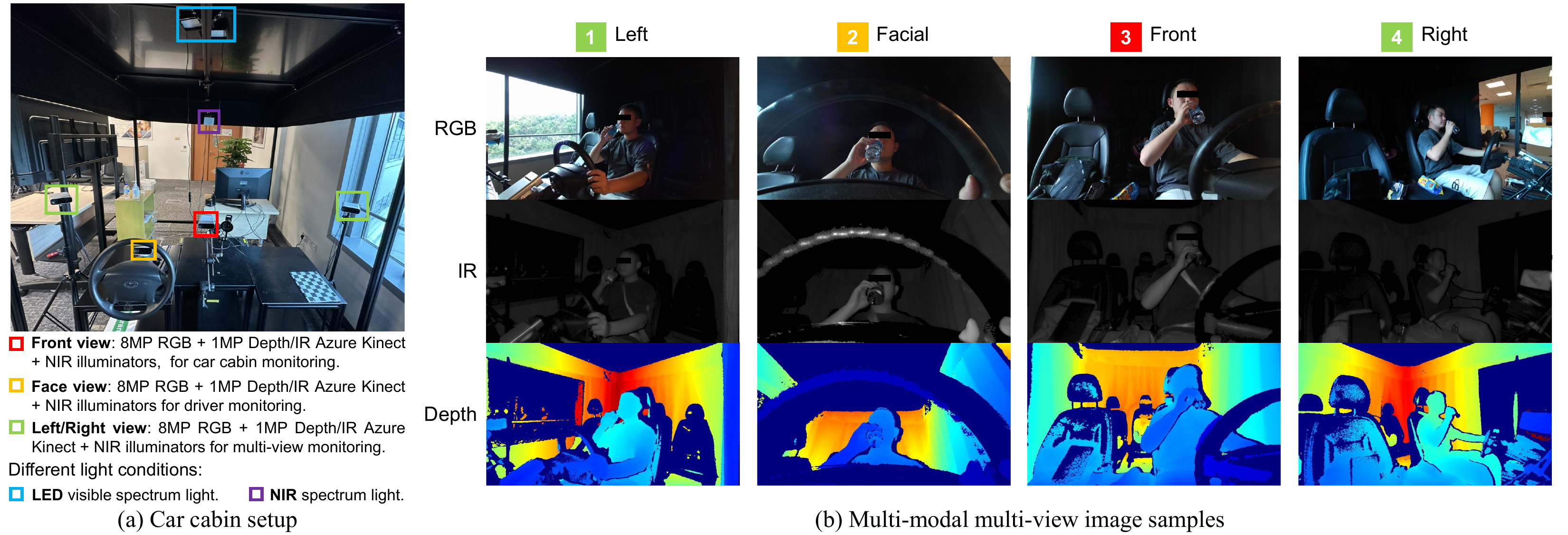}
\caption{Visualization and setup of the proposed DAOS dataset. 
    (a) Car cabin setup with four Azure Kinect devices (front, facial, left, right) capturing RGB, Depth/IR under LED and NIR illumination.  
    (b) Multimodal multiview image samples from the four camera views (Left, Facial, Front, Right) and three sensing modalities (RGB, IR, Depth).}
\label{fig:setup}\vspace{-5pt}
\end{figure*}

In this section, we introduce a new multi-modal, multi-view, multi-task car cabin monitoring dataset, including its collection, annotation and statistics.
\vspace{-5pt}
\subsection{Dataset Collection}
To safely simulate in-cabin scenarios for action recognition, a dedicated car cabin simulator was constructed, allowing controlled data collection while safeguarding participants. The simulator setup incorporated four Azure Kinect cameras configured to capture RGB data at a resolution of 1920×1080, IR and depth data at 1024×1024, all at a frame rate of 15 fps. A curtain was used to mimic day and night scenarios, with additional LED light to simulate the lighting environment.

Our dataset includes recordings from four perspectives (front, face, left, and right) to ensure comprehensive coverage of in-cabin actions. A wide range of actions was captured, and organized into 12 coarse classes and 36 fine-grained classes, enhancing the diversity of our dataset.

Figure~\ref{fig:setup} showcases the dataset's multi-perspective and multi-modal characteristics. The left panel presents a top-down view of the data collection setup, illustrating four camera perspectives (\textit{Left}, \textit{Face}, \textit{Front}, and \textit{Right}) and their placement relative to the car cabin. These cameras captured RGB, IR, and Depth modalities simultaneously. The steering wheel and face cameras were fixed, while the cabin layout simulated a realistic vehicle with two front seats and three rear seats.

The right panel of Figure~\ref{fig:setup} shows example samples for the action \textit{drinking}. The samples are arranged into a grid where:
\begin{itemize}
    \item \textbf{Rows Represent Modalities}: The first, second, and third rows correspond to RGB, IR, and Depth modalities, respectively.
    \item \textbf{Columns Represent Perspectives}: The four columns correspond to the four camera perspectives (\textit{Left}, \textit{Face}, \textit{Front}, and \textit{Right}).
\end{itemize}

These samples highlight the dataset's comprehensive coverage of perspectives and modalities, providing diverse visual inputs for training robust models. Such multi-perspective and multi-modal setups enhance the dataset's applicability to real-world scenarios and complex tasks like object-augmented action recognition modeling.

A total of 44 participants (10 female, 34 male) contributed to the data collection. Each participant recorded two sessions of identical actions, with the sequence of actions, positions of objects, and lighting conditions varying across sessions. This variability, combined with individual differences in executing specific actions (e.g., how they pick up objects, smoke, or use a phone), increases the dataset’s variety. Data collection sessions covered various times of the day and weather conditions, including sunny, cloudy, and rainy scenarios, capturing actions at the morning, noon, and afternoon intervals. This setup provides a robust dataset representing varied in-cabin scenarios, enhancing generalizability and supporting diverse action recognition applications.

\vspace{-5pt}
\subsection{Preprocessing and Annotation}
The preprocessing of recorded data involved aligning and standardizing information across multiple modalities and camera perspectives. To achieve precise temporal synchronization between the four perspectives, Azure’s external connection cables were used to synchronize across devices. For spatial alignment between the RGB and IR/depth views, each recording began with participants displaying a color checkerboard in front of the cameras. This setup provided consistent color and corner point references for calibration, allowing subsequent alignment adjustments based on the color and geometry of the checkerboard. Furthermore, IR and depth data were converted from 16 to 8 bits, optimizing compatibility and reducing data storage demands.

\textbf{Object Annotation:} Following preprocessing, we conducted a structured manual annotation process. Object annotation was performed using CVAT (Computer Vision Annotation Tool), where keyframes with significant changes in object state were identified and annotated. The intermediate frames were then interpolated to smoothly propagate the annotations, thereby minimizing manual workload while maintaining the accuracy of the annotations across sequences.

\textbf{Action Annotation:} For action annotation, we documented the start and end frames of each action in a structured table, allowing us to segment the data into three-second clips based on the action timestamps. This approach provides consistent action sequences for training and evaluation, enhancing the dataset’s usability in action recognition models.

\vspace{-10pt}
\subsection{Dataset Statistics}
Our DAOS dataset comprises 74 hours of synchronized multi-modal data captured from four calibrated perspectives, totaling 9,800 annotated video segments. As Table \ref{tab:data_attribute} demonstrates, this represents a significant advancement over existing driver monitoring datasets in three key dimensions:

\textbf{Scale and Diversity:} With 44 participants (10 female, 34 male), DAOS covers 7 more subjects than the largest comparable dataset DMD \cite{ortega2022challenges}. The inclusion of varied lighting conditions (day/night/rain) and 36 fine-grained action categories ensures comprehensive coverage of in-cabin scenarios. As shown in Figure~\ref{fig:category}, we capture the videos across different weather and lighting conditions. A shading curtain was used to mimic day and night scenarios, with additional LED lighting to simulate the night lighting environment. The dataset's technical superiority is further evidenced by its multi-view coverage—4 synchronized perspectives versus Drive\&Act's 6 unsynchronized views and DMD's 3 views. 

\textbf{Multimodal Richness:} Unlike 78\% of existing datasets limited to single or two modalities such as DriverMVT \cite{DriverMVT}, DAOS provides synchronized RGB, IR, and depth streams. Our depth resolution (1024×1024) surpasses Drive\&Act's 512×424, enabling precise 3D localization of objects and relations.

\textbf{Object-Centric Annotation:} DAOS introduces 15 object categories with 2.58 million annotated instances. Our DAOS dataset introduces pioneering object-centric annotations with label and location metadata, the first among in-cabin monitoring datasets. As Figure \ref{fig:category} illustrates, 64\% of fine-grained actions directly correlate with specific object states, enabling joint action-object reasoning.




\begin{figure*}[h]
\centering
\includegraphics[width=\linewidth]{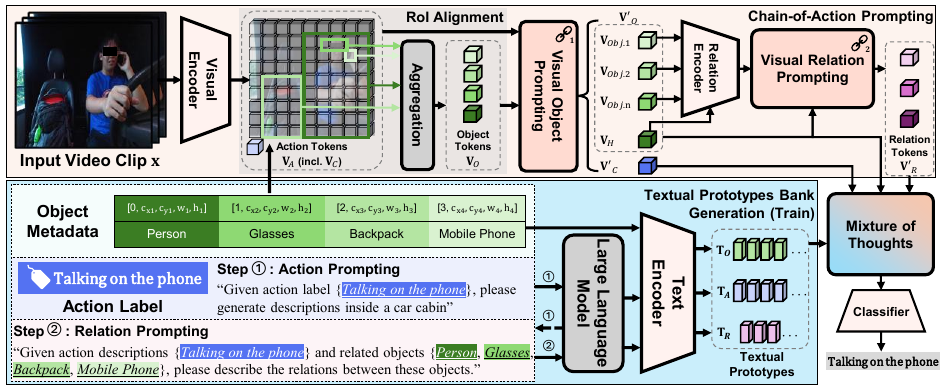}
\caption{An overview of the proposed AOR-Net framework. The model employs a Chain-of-action Prompting structure to sequentially extract and refine action-level, object-level, and relation-level features using attention in Object Prompting and Relation Prompting. These features are then aligned with their corresponding textual embeddings using a differentiable one-hot alignment. The Mixture of Thoughts (MoT) module dynamically adjusts the weights of these aligned features to produce the final action classification.}
\label{fig:framework}\vspace{-10pt}
\end{figure*}

To ensure rigorous evaluation, we adopt a participant-wise split (Train:32, Val:6, Test:6) that prevents data leakage. This partitioning strategy, combined with the scale and annotation density of DAOS, supports novel research directions in cross-modal fusion and occlusion reasoning through multi-view alignment, which are unrealizable with existing in-cabin datasets. Moreover, the inclusion of extensive object annotations enables support for multi-task learning and multi-modal information fusion studies, positioning our dataset as a versatile resource for in-cabin action recognition research.

\section{Method}\label{Method}

\subsection{Preliminaries and Motivation}
Recognizing driver actions in the cabin environment poses two key challenges: many actions are visually similar and can only be disambiguated by subtle object interactions, and the importance of object versus relation cues varies widely across action classes, making fixed-feature fusion suboptimal. To tackle these issues, we propose AOR-Net, built on Open-VCLIP~\cite{pmlr-v202-weng23b}, which comprises a Chain-of-action prompting (CoA) framework and a Mixture of Thoughts (MoT) module. CoA, inspired by chain-of-thought prompting~\cite{wei2022chain}, decomposes recognition into three sequential reasoning steps—extracting coarse action tokens, refining them with object-centric features via RoIAlign and an MLP, and then incorporating LLM-generated relation prompts to model human–object interactions (see Figure~\ref{fig:framework}). MoT addresses the drawbacks of static weighting by dynamically aligning and weighting action-, object-, and relation-level embeddings, allowing the model to emphasize the most informative cues for each action class and thereby enhance generalization. The overall AOR-Net architecture is shown in Figure~\ref{fig:framework}.

\subsubsection{Object-augmented Action Recognition (OAR)}
is designed to enhance video-based action recognition by explicitly incorporating object-level context and relations. Formally, given a video clip \(\mathbf{x}\) and its per-frame object detections \(\{(b_i, c_i)\}_{i=1}^N\), where each \(b_i\in\mathbb{R}^4\) is a bounding box and \(c_i\) an object class, the goal is to predict the action label \(y\). Traditional models extract global spatio-temporal features but often fail to distinguish actions with similar motion patterns. By leveraging object appearance and location, object-augmented methods can disambiguate fine-grained actions and improve recognition robustness.

\subsubsection{Relation-prompted OAR}
Recognizing driver actions in the cabin environment poses two key challenges: many actions are visually similar and can only be disambiguated by subtle object interactions, and the importance of object versus relation cues varies widely across action classes, making fixed-feature fusion suboptimal. To tackle these issues, we propose AOR-Net, built on Open-VCLIP~\cite{pmlr-v202-weng23b}, which comprises a Chain-of-Action Prompting (CoA) framework and a Mixture of Thoughts (MoT) module. CoA, inspired by chain-of-thought prompting~\cite{wei2022chain}, decomposes recognition into three sequential reasoning steps: extracting coarse action tokens, refining them with object-centric features via RoIAlign and an MLP, and then incorporating LLM-generated relation prompts to model human–object interactions. MoT addresses the drawbacks of static weighting by dynamically aligning and weighting action-, object-, and relation-level embeddings, allowing the model to emphasize the most informative cues for each action class and thereby enhance generalization. The overall AOR-Net architecture is shown in Figure~\ref{fig:framework}.

\vspace{-2pt}
\vspace{-5pt}
\subsection{Chain-of-action Prompting}
\label{sec:coa_prompt}


The Chain-of-action (CoA) prompting decomposes action understanding into three hierarchical stages: \textbf{action-level}, \textbf{object-level}, and \textbf{relation-level}. 
Inspired by the chain-of-thought prompting paradigm~\cite{wei2022chain}, which progressively reasons through intermediate steps, our AOR-Net models visual understanding as a sequential reasoning process. 
The model first captures coarse action cues from raw frames, then injects object-level semantics, and finally refines relational representations to infer detailed human–object interactions.

\textbf{Action-level:}  
Given an input video clip \( \mathbf{x} \), the visual encoder \( f_{\theta_V} \) of CLIP extracts a sequence of visual tokens 
\( \mathbf{V}_A \in \mathbb{R}^{(THW+1) \times d} \), 
which includes a global class token \( \mathbf{V}_C \in \mathbb{R}^{d} \). 
Here, \( T, H, W \) denote the temporal, height, and width dimensions of the token grid, and \( d \) represents the embedding dimension. 
These action-level tokens serve as the base representation summarizing the overall spatio-temporal dynamics of the video.

\textbf{Object-level:}  
To enhance action representations with localized semantics, object cues are introduced at this stage. 
Given a set of detected bounding boxes 
\( \mathfrak{B} \in \mathbb{R}^{TO \times 4} \) 
for \( O \) objects across \( T \) frames, RoIAlign~\cite{he2017mask} is applied to the spatial tokens of the visual feature map \( \mathbf{V}_A \) to extract region-specific features for each object:
\begin{equation}
    \mathbf{v}_{o}^{t} = \text{RoIAlign}(\mathbf{V}_A^t, \mathfrak{B}_{o}^{t}),
\end{equation}
where \( \mathbf{V}_A^t \) denotes the spatial tokens at time step \( t \), and \( \mathbf{v}_{o}^{t} \) is the corresponding object patch feature.  
Each cropped feature is then processed by a two-layer Multi-Layer Perceptron (MLP) followed by max pooling along the temporal dimension:
\begin{equation}
    \mathbf{V}_O = \text{MaxPool}_t(\text{MLP}_2(\mathbf{v}_{o}^{t})) \in \mathbb{R}^{TO \times d},
\end{equation}
where \( d \) denotes the shared embedding dimension between all feature levels.  
This operation extracts fine-grained spatial and temporal patterns for each object while maintaining temporal compactness.

To enable bidirectional information exchange between the global action context and local object cues, 
action and object tokens are concatenated and refined through a multi-head self-attention mechanism (MHSA):
{\setlength{\abovedisplayskip}{5pt}
 \setlength{\belowdisplayskip}{5pt}
\begin{equation}
    [\mathbf{V}_A'; \mathbf{V}_O'] = \text{MHSA}([\mathbf{V}_A; \mathbf{V}_O]),
\end{equation}}
where \( \mathbf{V}_A' \) and \( \mathbf{V}_O' \) are the updated action and object embeddings, respectively.  
The MHSA operation allows each token to attend to others across the joint token set, thereby introducing global receptive fields.  

As a result, object tokens gain contextual awareness of the overall scene and co-occurring actions, while action tokens learn to focus adaptively on the most relevant objects within the cabin environment. This interaction step effectively bridges coarse global motion with localized object dynamics, forming a refined representation for subsequent relational reasoning.

\textbf{Relation-level:}  
At this stage, human–object interactions are modeled explicitly to capture relational semantics crucial for understanding driver intent.  
The refined object tokens \( \mathbf{V}_O' \) are partitioned into human tokens \( \mathbf{V}_H \) and non-human object tokens \( \mathbf{V}_{Obj} \) according to their class indices.  
All possible human–object pairs are enumerated, and each pair’s concatenated feature is passed through a five-layer Relation Encoder implemented as an MLP:
\begin{equation}
    \mathbf{V}_R = \text{MLP}_5([\mathbf{V}_H; \mathbf{V}_{Obj}]),
\end{equation}
where each hidden layer has 1024 units and nonlinear activation, and the final layer projects the output back to the shared embedding dimension \( d \).  
This step learns compact relation embeddings that capture interaction intensity, spatial proximity, and temporal co-occurrence between the driver and cabin objects.

To further refine these relation embeddings and enhance contextual coherence, we apply a multi-head cross-attention (MHCA) mechanism:
\begin{equation}
    \mathbf{V}_R' = \text{MHCA}(\mathbf{Q}=\mathbf{V}_R,\; \mathbf{K}=\mathbf{V}_O',\; \mathbf{V}=\mathbf{V}_O').
\end{equation}
Here, the relation tokens \( \mathbf{V}_R \) act as the query set, while the object tokens \( \mathbf{V}_O' \) serve as both keys and values.  

This design allows relation tokens to selectively attend to object-level information, integrating cues about neighboring entities, occlusion patterns, and spatial configurations. Through this attention refinement, each relation token develops an adaptive understanding of the surrounding object context, enhancing discriminability for similar actions (e.g., \textit{“picking up a phone”} vs. \textit{“holding a phone”}) that depend on subtle relational cues.

Overall, this hierarchical prompting chain mechanism progressively transforms the representation from coarse spatio-temporal action features to semantically grounded object features, and finally to explicit human–object relational embeddings.  
After these stages, the enhanced visual features \( \{\mathbf{V}_A', \mathbf{V}_O', \mathbf{V}_R'\} \) jointly encode both global scene dynamics and fine-grained relational semantics, establishing a robust foundation for subsequent cross-modal alignment in AOR-Net.

\vspace{-10pt}
\subsection{Textual Prototype Bank Generation}
\label{sec:model_details}

To construct a unified textual prototype bank aligned with multi-level visual representations, we employ the GPT-4o model to generate descriptive text for both actions and their associated human–object relations within the car cabin environment. 
The process consists of two main stages: (1) \textbf{action description generation} and (2) \textbf{relation description generation}, followed by an expert verification and refinement loop to ensure contextual accuracy and descriptive diversity. 
All validated descriptions are then encoded by the CLIP text encoder \( f_{\theta_T} \) to produce hierarchical textual features for each reasoning level, forming the final prototype bank \( \{\mathbf{T}_A, \mathbf{T}_O, \mathbf{T}_R\} \).

\begin{algorithm}[t]
\caption{Textual Prototype Bank Generation}
\label{alg:textbank}
\begin{algorithmic}[1]
\REQUIRE Action label set $\mathcal{L}_A=\{l_A^1,\ldots,l_A^{N_A}\}$, object label set $\mathcal{L}_O=\{l_O^1,\ldots,l_O^{N_O}\}$
\ENSURE Textual prototype bank $\{\mathbf{T}_A, \mathbf{T}_O, \mathbf{T}_R\}$

\STATE Initialize $\mathbf{T}_A, \mathbf{T}_O, \mathbf{T}_R \leftarrow \emptyset$
\STATE \textbf{Stage 1: Action Description Generation}
\FOR{$l_A \in \mathcal{L}_A$}
    \STATE $\text{desc}_A \leftarrow \text{GPT4o}(l_A, \text{prompt}_A)$
    \IF{$\neg \text{Valid}(\text{desc}_A)$}
        \STATE $\text{prompt}_A \leftarrow \text{RefinePrompt}(\text{prompt}_A)$
        \STATE $\text{desc}_A \leftarrow \text{GPT4o}(l_A, \text{prompt}_A)$
    \ENDIF
    \STATE $\mathbf{T}_A \leftarrow f_{\theta_T}(\text{tokenize}(\text{desc}_A))$
\ENDFOR

\STATE \textbf{Stage 2: Relation Description Generation}
\FOR{$l_A \in \mathcal{L}_A$}
    \STATE $\mathcal{O}(l_A) \leftarrow \text{GetObjects}(l_A, \mathcal{L}_O)$
    \STATE $\text{desc}_R \leftarrow \text{GPT4o}(l_A, \mathcal{O}(l_A), \text{prompt}_R)$
    \IF{$\neg \text{Valid}(\text{desc}_R)$}
        \STATE $\text{prompt}_R \leftarrow \text{RefinePrompt}(\text{prompt}_R)$
        \STATE $\text{desc}_R \leftarrow \text{GPT4o}(l_A, \mathcal{O}(l_A), \text{prompt}_R)$
    \ENDIF
    \STATE $\mathbf{T}_R \leftarrow f_{\theta_T}(\text{tokenize}(\text{desc}_R))$
\ENDFOR
\STATE $\text{ExportToJSON}(\text{desc}_A, \text{desc}_R)$

\STATE \textbf{Stage 3: Object Prototype Encoding}
\FOR{$l_O \in \mathcal{L}_O$}
    \STATE $\mathbf{T}_O \leftarrow f_{\theta_T}(\text{tokenize}(l_O))$
\ENDFOR

\STATE \textbf{Stage 4: Prototype Bank Assembly}
\STATE $\mathcal{B}_T \leftarrow \{\mathbf{T}_A, \mathbf{T}_O, \mathbf{T}_R\}$
\RETURN $\mathcal{B}_T$
\end{algorithmic}
\end{algorithm}

\subsubsection{Action Description Generation}
For each action label \( l_A \in \mathcal{L}_A \), GPT-4o generates textual descriptions using a cabin-aware prompt:
\begin{quote}
\textit{“Given action label \{class name\}, please generate descriptions inside a car cabin.”}
\end{quote}
The generated set of descriptions \( \text{desc}_A = \{d_A^1, \ldots, d_A^K\} \) is evaluated by the expert verifier for contextual accuracy and diversity. 
If \( \text{desc}_A \) fails the validation function \( \text{Valid}(\cdot) \), the prompt \( \text{prompt}_A \) is refined via \( \text{RefinePrompt}(\cdot) \) and regenerated.
The final accepted descriptions are tokenized and encoded by the CLIP text encoder:
{\setlength{\abovedisplayskip}{4pt}
 \setlength{\belowdisplayskip}{4pt}
\begin{equation}
    \mathbf{T}_A = f_{\theta_T}(\text{tokenize}(\text{desc}_A)) \in \mathbb{R}^{N_A \times d}.
\end{equation}}

\subsubsection{Relation Description Generation}
For each action \( l_A \), the corresponding object set \( \mathcal{O}(l_A) \subseteq \mathcal{L}_O \) is identified, and GPT-4o generates relational sentences using the prompt:
\begin{quote}
\textit{“Given action descriptions \{action description\} and related objects \{object names\}, please describe the relations between these objects.”}
\end{quote}
The generated relation descriptions \( \text{desc}_R = \{d_R^1, \ldots, d_R^M\} \) are validated by the expert verifier; if the coverage or consistency is insufficient, a refined prompt \( \text{prompt}_R' \) is used to regenerate. 
The approved descriptions are tokenized and encoded as relation-level textual prototypes:
{\setlength{\abovedisplayskip}{4pt}
 \setlength{\belowdisplayskip}{4pt}
\begin{equation}
    \mathbf{T}_R = f_{\theta_T}(\text{tokenize}(\text{desc}_R)) \in \mathbb{R}^{N_R \times d}.
\end{equation}}

\subsubsection{Object Prototype Encoding and Bank Assembly}
All object category labels \( l_O \in \mathcal{L}_O \) are encoded using the same text encoder:
{\setlength{\abovedisplayskip}{4pt}
 \setlength{\belowdisplayskip}{4pt}
\begin{equation}
    \mathbf{T}_O = f_{\theta_T}(\text{tokenize}(l_O)) \in \mathbb{R}^{N_O \times d}.
\end{equation}}
Finally, all encoded features are aggregated into a structured textual prototype bank:
{\setlength{\abovedisplayskip}{4pt}
 \setlength{\belowdisplayskip}{4pt}
\begin{equation}
    \mathcal{B}_T = \{\mathbf{T}_A, \mathbf{T}_O, \mathbf{T}_R\}.
\end{equation}}

This unified textual prototype bank provides multi-level semantic anchors for cross-modal alignment in the MoT module, enabling joint reasoning over actions, objects, and relations. The iterative generation and expert verification loop ensures that all textual prototypes are accurate, diverse, and contextually aligned with the DAOS dataset.
\vspace{-12pt}

\subsection{Mixture of Thoughts (MoT)}
\begin{figure}[t]
\centering
\includegraphics[width=\linewidth]{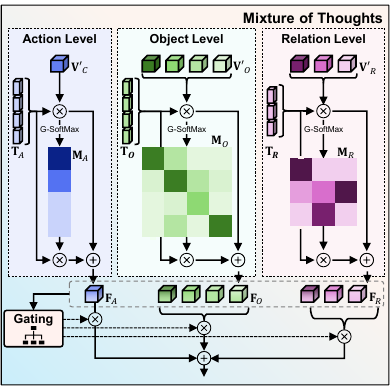}
\caption{Details of the Mixture of Thoughts module. The MoT module aligns these multi-level visual and textual features via a differentiable one-hot trick and dynamically fuses them to produce the final action prediction.}
\label{fig:mot}\vspace{-10pt}
\end{figure}

The in-cabin dataset contains various actions, some involving objects and others not. The relevance of object-level and relation-level information to the final classification affects the model's performance. Traditionally, a fixed weighted sum of the three levels is used to produce the final prediction. However, static weighting lacks adaptability, as object and relation importance vary significantly across action classes. To address this, we propose the MoT module, which dynamically adjusts weights based on aligned action, object, and relation features to enhance generalization. An illustration of the MoT module is shown in Figure~\ref{fig:mot}.

First, we compute the similarity matrices between the updated visual features and their corresponding textual features for each level: $\mathbf{M}_l = \mathbf{V}_l' \cdot \mathbf{T}_l^\top, \quad l \in \{A, O, R\}$,
where \( \mathbf{V}_C', \mathbf{V}_O', \mathbf{V}_R' \) are the updated visual features from the action, object, and relation levels, and \( \mathbf{T}_A, \mathbf{T}_O, \mathbf{T}_R \) are the corresponding textual features. We use $\mathbf{V}_C'$ to replace $\mathbf{V}_A'$ as an action class token.

To eliminate ambiguity in alignment, we implement the differentiable one-hot trick~\cite{chen2024align} using Gumbel Softmax for each level, as
{\setlength{\abovedisplayskip}{5pt}
 \setlength{\belowdisplayskip}{5pt}
\begin{align}
    &\mathbf{\hat{M}}_l = \text{one-hot}(\mathbf{M}_{l,\text{argmax}}) + \mathbf{M}_l \notag - \text{detach}(\mathbf{M}_l), \\
    & l \in \{A, O, R\}. \label{eq:one_hot}
\end{align}}

The aligned textual features are then obtained and combined with their respective visual features: $\mathbf{F}_l = \mathbf{V}_l' + \text{MLP}(\mathbf{\hat{M}}_l \cdot \mathbf{T}_l)$, where $l \in \{A, O, R\}$.

The combined features from all levels are concatenated to form the input to the MoT module. To generate dynamic weights, the MoT module flattens combined features and applies a multi-layer perceptron (MLP) followed by a softmax function shown as $\mathbf{W} = \sigma(\text{MLP}(\text{flatten}([\mathbf{F}_A; \mathbf{F}_O; \mathbf{F}_R])))$.
where \( \mathbf{W} = [W_A, \mathbf{W}_O, \mathbf{W}_R] \in \mathbb{R}^{1 + O + R} \), with \( W_A \in \mathbb{R} \) being a scalar for the action token, \( \mathbf{W}_O \in \mathbb{R}^O \) providing weights for the object tokens, and \( \mathbf{W}_R \in \mathbb{R}^R \) providing weights for the relation tokens. The final feature representation is computed by combining the weighted features:
{\setlength{\abovedisplayskip}{5pt}
 \setlength{\belowdisplayskip}{5pt}
\begin{equation}
    \mathbf{A}_{\text{final}} = W_A \cdot \mathbf{F}_A + \sum_{i=1}^{O} W_{O,i} \cdot \mathbf{F}_{O,i} + \sum_{j=1}^{R} W_{R,j} \cdot \mathbf{F}_{R,j}. \label{eq:afinal}
\end{equation}}

\begin{table*}[t]
\centering
\caption{Comparison of state-of-the-art methods on fine-grained and coarse levels of DAOS Dataset.}
\label{tab: DAOS}
\setlength{\abovecaptionskip}{5pt}
\begin{threeparttable}
  \resizebox{\linewidth}{!}{
    \begin{tblr}{
      colspec = {lcccccccccccc},
      row{1} = {c},
      cell{1}{1} = {r=3}{c},
      cell{1}{2} = {c=6}{},
      cell{1}{8} = {c=6}{},
      cell{2}{2} = {c=3}{},
      cell{2}{5} = {c=3}{},
      cell{2}{8} = {c=3}{},
      cell{2}{11} = {c=3}{},
      hline{1,13} = {-}{0.1em},
      hline{2,3,4,10} = {-}{},
      vline{5} = {-}{},
      vline{8} = {-}{},
      vline{11} = {-}{},
    }
    Methods & \SetCell[c=6]{} Validation & & & & & & \SetCell[c=6]{} Test & & & & & \\
            & \SetCell[c=3]{} Fine-grained & & & \SetCell[c=3]{} Coarse & & & \SetCell[c=3]{} Fine-grained & & & \SetCell[c=3]{} Coarse & & \\
            & Top-1 & Top-5 & Mean-1 & Top-1 & Top-5 & Mean-1 & Top-1 & Top-5 & Mean-1 & Top-1 & Top-5 & Mean-1 \\
    ViViT-B \cite{arnab2021vivit} & 42.36  & 74.12 & 28.90 & 48.46 & 87.96 & 39.22 & 39.47 & 69.43 & 30.35 & 43.24 & 85.14 & 36.22 \\
    TimeSFormer-B \cite{bertasius2021timesformer} & 43.11  & 75.53 & 29.15 & 54.55 & 88.03 & 48.93 & 40.21 & 70.28 & 28.42 & 49.64 & 88.38 & 44.37 \\ 
    ORViT \cite{herzig2022object} & 51.05  & 79.35 & 39.33 & 57.74 & 88.80 & 52.66 & 50.15 & 78.38 & 37.24 & 53.05 & 89.19 & 51.98 \\
    VideoMAE \cite{tong2022videomae} & 52.55  & 80.86 & 42.16 & 60.00 & 89.70 & 52.50 & 50.98 & 79.35 & 39.26 & 57.19 & 88.25 & 50.38 \\
    Open-VCLIP \cite{pmlr-v202-weng23b} & 53.76  & 81.03 & 42.96 & 63.74 & 90.85 & 58.29 & 53.30 & 78.30 & 42.41 & 62.17 & 89.68 & 57.70 \\
    AOR-Net (Ours) & \textbf{55.51} & \textbf{82.50} & \textbf{45.39} & \textbf{65.73} & \textbf{92.08} & \textbf{62.60} & \textbf{55.08} & \textbf{81.99} & \textbf{44.97} & \textbf{64.49} & \textbf{90.15} & \textbf{61.68} \\ 
    Roitberg \textit{et al.}\textsuperscript{$\dagger$} \cite{roitberg2022comparative} & 58.44 & 86.45 & 50.01 & 62.50 & 92.92 & 61.22 & 54.43 & 78.17 & 42.03 & 60.66 & 89.03 & 56.87 \\
    Open-VCLIP\textsuperscript{$\dagger$} \cite{pmlr-v202-weng23b} & 61.98 & 86.15 & 51.26 & 67.29 & 91.46 & 63.30 & 57.07 & 82.07 & 43.96 & 67.09 & 89.14 & 63.12 \\
    AOR-Net (Ours)\textsuperscript{$\dagger$} & \textbf{63.75} & \textbf{87.50} & \textbf{55.00} & \textbf{73.96} & \textbf{94.90} & \textbf{72.16} & \textbf{61.39} & \textbf{82.28} & \textbf{46.72} & \textbf{70.25} & \textbf{89.24} & \textbf{64.41} \\
    \end{tblr}
  }
\begin{tablenotes}
\small
\item[$\dagger$] Results obtained using multi-modal inputs (RGB+IR+Depth).
\end{tablenotes}
\end{threeparttable}\vspace{-8pt} 
\end{table*}

The final prediction is produced through a fully connected layer applied to \( \mathbf{A}_{\text{final}} \), as $\hat{y} = \text{FC}(\mathbf{A}_{\text{final}})$.
\vspace{-5pt}
\subsection{Training and Inference}
\textbf{Training Stage.} During training, our method takes video clips, action descriptions, object labels, and relation descriptions as inputs. To prevent data leakage and avoid introducing action-specific objects and relations directly into the input, we input the entire dataset's action descriptions, object labels, and relation descriptions into the model. Both the visual and textual inputs are processed separately through the CLIP model and the CoA prompting module, resulting in updated visual features \( \mathbf{V}_A' \), \( \mathbf{V}_O' \), \( \mathbf{V}_R' \) and textual features \( \mathbf{T}_A \), \( \mathbf{T}_O \), \( \mathbf{T}_R \). These features are then aligned and combined in the MoT module, with the final prediction obtained via a fully connected (FC) layer.

The loss function used is a cross-entropy loss. For a given sample's final feature representation \( \mathbf{A}_{\text{final}}^n \), the loss function is formulated as: 
{\setlength{\abovedisplayskip}{3pt}
 \setlength{\belowdisplayskip}{5pt}
\begin{align}
\hat{y} = \text{FC}(\mathbf{A}_{\text{final}}) \quad
    \mathcal{L} = -\frac{1}{N} \sum_{n=1}^{N} y_{n} \log(\hat{y}_{n}),
\label{eq:loss_function}
\end{align}}

where \( N \) is the number of video samples, \( y_n \) is the ground truth label for \( n \)-th video, and \( \hat{y}_n \) is the predicted probability.

\textbf{Inference Stage.} During inference, the input process remains consistent with the training stage. The model receives the textual information for the entire dataset and generates predictions through the CoA and MoT modules, resulting in the final action classification.
\section{Experiments}\label{Exp}
\subsection{Implementation Details}\label{implement}

For training stage, we utilize the Open-VCLIP \cite{pmlr-v202-weng23b} model pre-trained on Kinetics-400 \cite{kay2017kinetics} as our backbone. We first truncate the original videos into small clips with a length of three seconds in the preprocessing process according to the video category. Then temporally sample the clips to collect 8 frames randomly, and resize the video to 224x224 followed by RandomFlip operation. The token embedding size is (2, 16, 16), and the feature dimension is 768, which aligns with Open-VCLIP. We adopt GPT-4o to enrich textual information for actions and relations. For multimodal scenarios, our AOR-Net model encodes the data of the three modalities separately and uses the MoT module to integrate and select appropriate information for prediction. Unless otherwise stated, the maximum number of objects is set to 6 by default.

All single-modality experiments are conducted using 4 NVIDIA A5000 GPUs with a batch size of 16. For multi-modal experiments, the batch size is set to 4 to train the combined models end-to-end. The AdamW optimizer is utilized with a step learning rate of [10e-4, 10e-5, 10e-6] at [0, 15, 25] epochs. We train the model for 30 epochs in total using label-smoothing cross-entropy loss.

\vspace{-5pt}
\subsection{Action Recognition Results}\label{results}
We follow previous works \cite{martin2019drive,lin2024multi} and report three standard metrics: Top-1 Accuracy (Top-1), Top-5 Accuracy (Top-5), and Mean-1 Accuracy (Mean-1). 
Mean-1 represents the mean of per-class Top-1 accuracy, offering a more reliable measure under class-imbalanced conditions. 

Table~\ref{tab: DAOS} summarizes the results on both validation and test sets for the fine-grained and coarse-grained levels of the DAOS dataset. 
For fair comparison, we evaluate our AOR-Net against state-of-the-art (SOTA) transformer-based models and object-augmented methods discussed in Section~\ref{object_augment}. 
For multimodal evaluation, we employ RGB, IR, and Depth modalities, training three CoA branches individually and fusing them through the MoT module. 
Likewise, three Open-VCLIP \cite{pmlr-v202-weng23b} models are fine-tuned and fused in the same manner to ensure parity.

On the \textbf{validation set}, AOR-Net achieves 55.51\% Top-1 and 45.39\% Mean-1 accuracy for fine-grained actions, outperforming ViViT-B and TimeSFormer-B by over 12\% and 10\% in Top-1 respectively. 
Compared with ORViT and VideoMAE, AOR-Net delivers consistent gains of +4.46\% and +2.96\% in Top-1, verifying the benefit of explicit relation modeling. 
On the coarse-grained level, it further achieves 65.73\% Top-1 and 62.60\% Mean-1, setting a new benchmark on DAOS. 
When extended to the multimodal configuration (RGB+IR+Depth, marked by $\dagger$), our model reaches 73.96\% coarse-grained Top-1 and 55.00\% fine-grained Mean-1, improving upon Open-VCLIP by 6.67\% and 3.74\%, respectively.

On the \textbf{test set}, AOR-Net maintains strong generalization, obtaining 55.08\% Top-1 and 44.97\% Mean-1 on the fine-grained level under the RGB setting, the best among all competing methods. 
Under multimodal fusion, AOR-Net achieves 61.39\% Top-1 and 46.72\% Mean-1 on the fine-grained task, and 70.25\% Top-1 and 64.41\% Mean-1 on the coarse-grained task. This surpasses the transformer-based baseline Open-VCLIP Top-1 by +4.32\% and +3.16\%, and remains consistently higher across all metrics compared with ORViT and VideoMAE. 
These improvements demonstrate that the proposed Chain-of-action (CoA) and Mixture of Thoughts (MoT) modules effectively capture fine-grained relational cues while maintaining robustness under multi-modal integration.

Overall, AOR-Net exhibits clear advantages in both single- and multi-modal settings. 
Its performance gains on the test set confirm the scalability and generalization of our relation-aware prompting and dynamic fusion framework, particularly in handling complex, object-driven driver behaviors in the DAOS dataset.

\vspace{-5pt}
\subsection{Ablation Study}\label{ablation}
\begin{figure}[t]
    \centering
    \includegraphics[width=\linewidth]{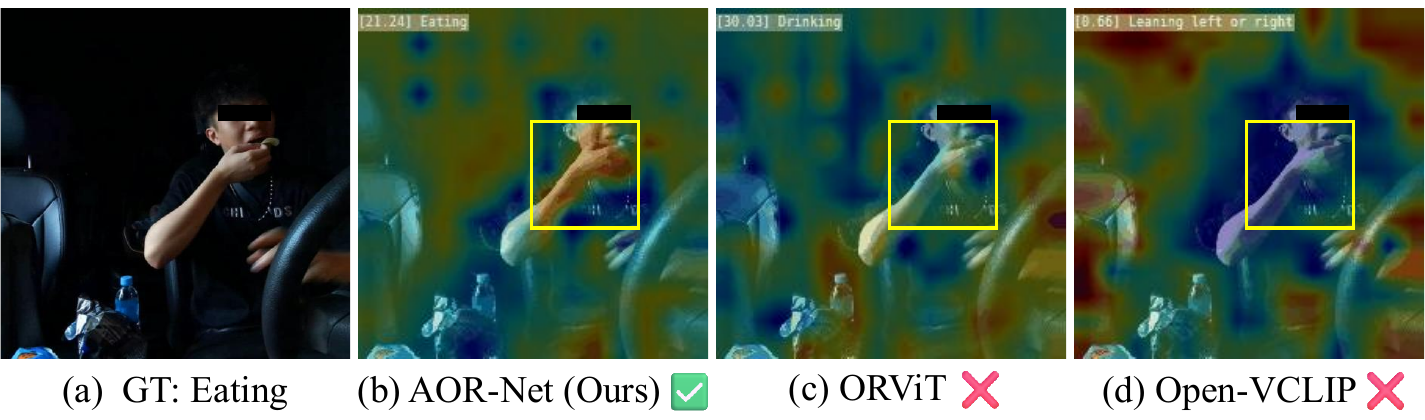}\vspace{-5pt}
    \caption{Comparative visualization of attention maps from different models in the last ViT layer. Check mark stands for correct prediction and cross mark means wrong predictions.}
    \label{fig:demo_supp}\vspace{-12pt}
\end{figure}

We systematically validate our architecture through three key experiments. We present visualizations of attention maps for better understanding as shown in Figure~\ref{fig:demo_supp}. Each component's contribution is analyzed as follows:

\textbf{Maximum object number in CoA Prompting.} Table \ref{tab: N_O} reveals the impact of maximum object retention per frame. Setting $\mathcal{O}$=6 achieves optimal performance with 55.51\% Top-1 and 45.39\% Mean-1 accuracy. Comparatively, $\mathcal{O}$=4 reduces Top-1 by 1.57\% due to insufficient object context, while $\mathcal{O}$=8/10 introduces noise from irrelevant objects, decreasing Top-1 by 0.52-0.75\%. This demonstrates our approach's effectiveness in balancing critical feature preservation with noise suppression.\vspace{-5pt}

\begin{table}[h]
\vspace{-3pt}
\centering
\setlength{\abovecaptionskip}{5pt}
\caption{Ablation study of different numbers of maximum objects per frame. }
\label{tab: N_O}
\resizebox{\linewidth}{!}{
\tiny
\begin{tblr}{
  rowsep = 1pt,   
  cells = {c},
  cell{1}{1} = {r=2}{},
  cell{1}{2} = {c=3}{},
  hline{1,7} = {-}{0.1em},
  hline{2} = {2-4}{},
  hline{3} = {-}{},
}
{\# Max Object\\ ($\mathcal{O}$) } & Metrics    &            &           \\
                         & Top-1      & Top-5      & Mean-1    \\
4                        & 53.94      & 81.62      & 43.11     \\
\textbf{6}                        & \textbf{55.51}      & \textbf{82.50}      & \textbf{45.39}     \\
8                        & 54.99      & 81.62      & 42.98     \\
10                       & 54.76      & 80.21      & 41.26     
\end{tblr}
}
\end{table}

\textbf{Effectiveness of different modules.} As shown in Table \ref{tab: module}, our baseline achieves 53.76\% Top-1 accuracy. The CoA Prompting alone boosts performance by 0.93\%, demonstrating its effectiveness in modeling dynamic object relations. When combined with Mixture of Thoughts (MoT), we observe further improvements of 1.75\% Top-1 and 2.43\% Mean-1, indicating complementary benefits between structured reasoning (CoA) and multi-perspective fusion (MoT).



\begin{table}[h]
\centering
\setlength{\abovecaptionskip}{5pt}
\caption{Ablation study of different settings of modules compared with baseline. CoA and MoT stand for CoA Prompting and Mixture of Thoughts, respectively.}
\label{tab: module}
\resizebox{\linewidth}{!}{
\tiny
\begin{tblr}{
  row{1} = {c},
  row{2} = {c},
  rowsep = 1pt,   
  cell{1}{1} = {r=2}{},
  cell{1}{2} = {c=3}{},
  cell{3}{2} = {c},
  cell{3}{3} = {c},
  cell{3}{4} = {c},
  cell{4}{2} = {c},
  cell{4}{3} = {c},
  cell{4}{4} = {c},
  cell{5}{2} = {c},
  cell{5}{3} = {c},
  cell{5}{4} = {c},
  hline{1,6} = {-}{0.1em},
  hline{2} = {2-4}{},
  hline{3} = {-}{},
}
Modules & Metrics        &                &                \\
        & Top-1          & Top-5          & Mean-1         \\
Base    & 53.76          & 81.03          & 42.96          \\
+ CoA   & 54.69          & 81.62          & 43.76          \\
+ CoA + MoT  & \textbf{55.51} & \textbf{82.50} & \textbf{45.39} 
\end{tblr}
}\vspace{-5pt}
\end{table}

\begin{figure}[t]
    \centering
    \includegraphics[width=\linewidth]{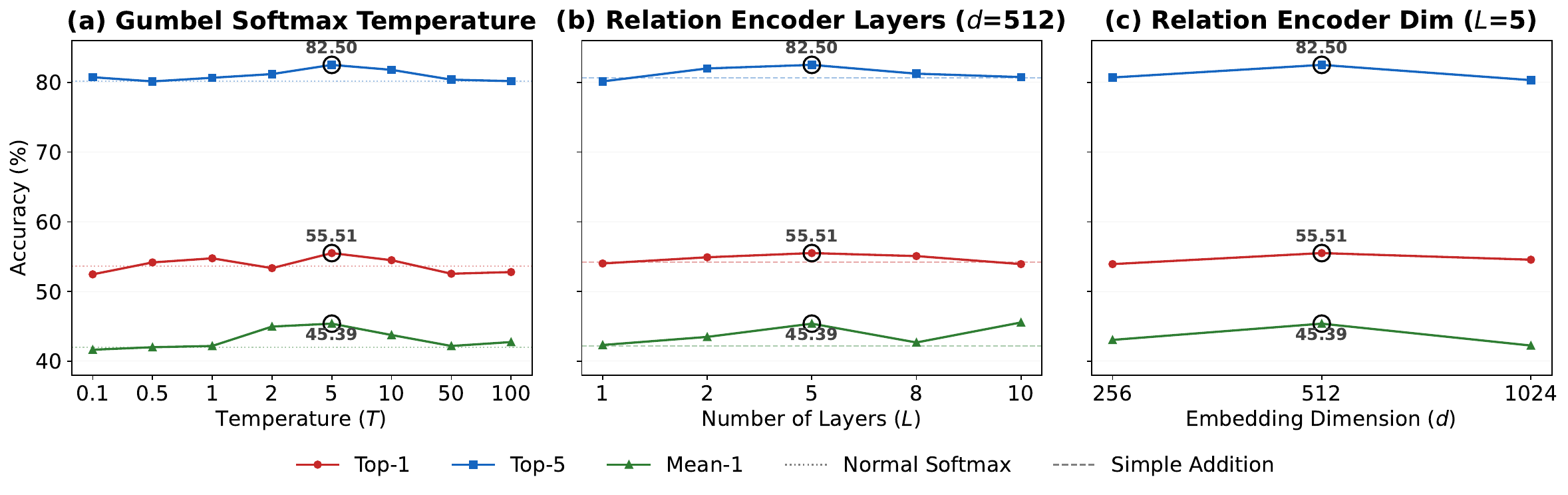}
    \caption{Ablation studies on key hyperparameters. (a)~Gumbel Softmax temperature $T$: dashed lines indicate the Normal Softmax baseline; performance peaks at $T{=}5$ (Top-1: 55.51\%, Top-5: 82.50\%, Mean-1: 45.39\%). (b)~Relation encoder depth with $d{=}512$: dashed lines indicate the Simple Addition baseline; the best accuracy is achieved at $L{=}5$. (c)~Relation encoder embedding dimension with $L{=}5$: $d{=}512$ yields the optimal balance across all three metrics.}
    \label{fig:ablation}\vspace{-10pt}
\end{figure}

\textbf{Gumbel Softmax Parameters Ablation.} Figure~\ref{fig:ablation} shows that incorporating Gumbel Softmax improves discrete alignment over the normal softmax baseline. Performance gradually increases as the temperature $T$ rises from 0.1 to 5.0, reaching the best result at $T{=}5.0$ (Top-1: 55.51\%, Top-5: 82.50\%, Mean-1: 45.39). Extremely small or large temperatures lead to unstable gradients or overly uniform sampling, confirming that a moderate range ($T{\approx}1$–$5$) provides the optimal trade-off between stochasticity and stability.

\textbf{Relation Encoder Parameters Ablation.} As summarized in Figure~\ref{fig:ablation}, deeper relation encoders consistently enhance performance up to five layers, after which accuracy saturates. The best configuration (L=5, $d{=}512$) achieves the highest Top-1 accuracy of 55.51\%, while shallower or excessively wide networks show marginal gains or even degradation. These results indicate that moderate depth and embedding dimension yield the most effective relational representation for action understanding.

\textbf{Cross-Dataset Ablation on Drive\&Act Subset.} We restrict comparisons to an RGB-only subset of Drive\&Act, for which we manually annotated dense object labels, as full-object annotation across the entire dataset was infeasible. As Table~\ref{tab: dna} shows, AOR-Net achieves 77.67\% Top-1 (versus 76.49\% for Open-VCLIP), validating the effectiveness of our CoA and MoT modules on cross-dataset evaluation. 

\begin{table}[t]
\centering
\setlength{\tabcolsep}{2pt}         
\caption{Comparison results on Drive\&Act \cite{martin2019drive} subset.}
\label{tab: dna}
\resizebox{\linewidth}{!}{
    \tiny
    \begin{tblr}{
      rowsep = 1pt,
      cells = {c},
      cell{1}{1} = {r=2}{},
      cell{1}{2} = {c=3}{},
      hline{1,6} = {-}{0.1em},
      hline{2} = {2-4}{},
      hline{3} = {-}{},
    }
    {Methods} & Metrics    &            &           \\
                            & Top-1      & Top-5      & Mean-1    \\
    ORViT \cite{herzig2022object}  & 73.75      & 93.47      & 62.15     \\
    Open-VCLIP \cite{pmlr-v202-weng23b} & 76.49      & 94.06      & 65.26     \\
    \textbf{AOR-Net}         & \textbf{77.67}      & \textbf{96.22}      & \textbf{66.90} 
    \end{tblr}
    }\vspace{-15pt}
\end{table}

\textbf{Multimodal fusion analysis.}
Table \ref{tab: multimodal} demonstrates progressive improvements through modality integration. Single modality shows RGB's advantage in Top-1 (55.51\%) versus Depth's strength in Mean-1 (48.62\%). Dual modality combinations achieve 4.80\% Top-1 gains, with RGB+Depth reaching 60.31\%. The full RGB+IR+Depth configuration peaks at 63.75\% Top-1, showing 13\% relative improvement over RGB alone. This 9.61\% Mean-1 gap confirms the effectiveness of our MoT module in combining multimodal cues.

\begin{table}[H]
\centering
\setlength{\tabcolsep}{2pt}         
\caption{Ablation study of single- and multi-modality model on fine-grained level data.}
\label{tab: multimodal}
\resizebox{\linewidth}{!}{
\tiny
\begin{tblr}{
  colspec = {lccc},
  row{1-2} = {c},
  rowsep = 1pt,   
  cell{1}{1} = {r=2}{},
  cell{1}{2} = {c=3}{},
  hline{1,10} = {-}{0.1em},
  hline{2} = {2-4}{},
  hline{3,6,9} = {-}{},
}
Modalities       & Metrics       &         &         \\
                 & Top-1         & Top-5   & Mean-1  \\
RGB              & 55.51         & 82.50   & 45.39   \\
IR               & 53.23         & 81.35   & 43.51   \\
Depth            & 54.58         & 83.54   & 48.62   \\
RGB+IR           & 61.15         & 85.73   & 50.89   \\
RGB+Depth        & 60.31         & 84.90   & 51.22   \\
IR+Depth         & 60.31         & 86.67   & 50.68   \\
RGB+IR+Depth     & 63.75         & 87.50   & 55.00   \\
\end{tblr}
}
\end{table}

\vspace{-8pt}
\textbf{Visualizations.} According to Figure~\ref{fig:demo_supp}, our model can predict most of the object-related classes correctly with explainable attention weights. Compared to existing object-augmented methods like ORViT \cite{herzig2022object}, AOR-Net can suppress the influence of irrelevant objects and background information. Unlike traditional action recognition methods (Open-VCLIP \cite{pmlr-v202-weng23b}) without the help of additional object information, AOR-Net can focus on related objects to predict the action correctly. 

\vspace{-10pt}
\section{Conclusion}
We propose a multi-modal Driver Action with Object Synergy (DAOS) dataset and Action-Object-Relation network (AOR-Net). DAOS consists of over 74 hours raw videos, including 12 coarse and 36 fine-grained action classes. Using our proposed Chain-of-action Prompting and Mixture of Thoughts module, information from different levels can be organized and selected properly. Through extensive experiments on DAOS, we observed that AOR-Net achieves state-of-the-art results and verified the effectiveness of our proposed modules and architecture. It is worth noting that our model can be adapted to any CLIP-based action recognition model. However, we have not invested in the generalization ability of the model and left it for follow-up research.


\ifCLASSOPTIONcaptionsoff
  \newpage
\fi



%
\bibliographystyle{IEEEtran}
\bibliography{./ref.bib}
\end{document}